\documentclass[sigconf]{acmart} %anonymous
\renewcommand\footnotetextcopyrightpermission[1]{} % removes footnote with conference information in first column
\usepackage{booktabs} % For formal tables
\usepackage{amsmath}

\usepackage{algorithm}
\usepackage[noend]{algpseudocode}
\usepackage{caption}
\usepackage{balance}
\usepackage[noabbrev,capitalise]{cleveref}
\usepackage{soul,color}

\setcopyright{rightsretained}
\copyrightyear{2019}
\acmYear{2019}
\setcopyright{acmcopyright}
\acmConference[GECCO '19 Companion]{Genetic and Evolutionary Computation Conference Companion}{July 13--17, 2019}{Prague, Czech Republic} \acmBooktitle{Genetic and Evolutionary Computation Conference Companion (GECCO '19 Companion), July 13--17, 2019, Prague, Czech Republic}
\acmPrice{15.00} \acmDOI{10.1145/3319619.3321939} \acmISBN{978-1-4503-6748-6/19/07}

\makeatletter
\if@ACM@anonymous
    \newcommand*{\ANONYMOUS}{}
\fi
\makeatother

\begin{document}
\title{Evaluating MAP-Elites on Constrained Optimization Problems}
%\titlenote{}
%\subtitle{}

\author{Stefano Fioravanzo}
% \authornote{}
% \orcid{}
\affiliation{%
  \institution{University of Trento}
  \streetaddress{via Sommarive 18}
  \city{Trento} 
  \state{Italy} 
  \postcode{38121}
}
\email{stefano.fioravanzo@studenti.unitn.it}
\additionalaffiliation{%
  \institution{Fondazione Bruno Kessler, fioravanzo@fbk.eu} %email added here to save one line in the authors section
  \streetaddress{Via Sommarive 18}
  \city{Trento} 
  \state{Italy} 
  \postcode{38123}
}
%\email{fioravanzo@fbk.eu}

\author{Giovanni Iacca}
% \authornote{}
\orcid{0000-0001-9723-1830}
\affiliation{%
  \institution{University of Trento}
  \streetaddress{Via Sommarive 18}
  \city{Trento} 
  \state{Italy} 
  \postcode{38123}
}
\email{giovanni.iacca@unitn.it}

%\renewcommand{\shortauthors}{S. Fioravanzo and G. Iacca}
\def\hyph{-\penalty0\hskip0pt\relax}

% ------------------------------------------------------------

\begin{abstract}
Constrained optimization problems are often characterized by multiple constraints that, in the practice, must be satisfied with different tolerance levels. While some constraints are hard and as such must be satisfied with zero-tolerance, others may be soft, such that non-zero violations are acceptable. Here, we evaluate the applicability of MAP-Elites to ``illuminate'' constrained search spaces by mapping them into feature spaces where each feature corresponds to a different constraint. On the one hand, MAP-Elites implicitly preserves diversity, thus allowing a good exploration of the search space. On the other hand, it provides an effective visualization that facilitates a better understanding of how constraint violations correlate with the objective function. We demonstrate the feasibility of this approach on a large set of benchmark problems, in various dimensionalities, and with different algorithmic configurations. As expected, numerical results show that a basic version of MAP-Elites cannot compete on all problems (especially those with equality constraints) with state-of-the-art algorithms that use gradient information or advanced constraint handling techniques. Nevertheless, it has a higher potential at finding constraint violations vs. objectives trade-offs and providing new problem information. As such, it could be used in the future as an effective building-block for designing new constrained optimization algorithms.
\end{abstract}



\maketitle

% ------------------------------------------------------------

\section{Introduction}\label{sec:intro}
Several real-world applications, for instance in engineering design, control systems and healthcare, can be described in the form of constrained continuous optimization problems, i.e. problems where a certain objective/cost function must be optimized within a certain search space, subject to some problem-dependent constraints. Without loss of generality, these problems can be formulated as:
\begin{eqnarray*}
	\underset{\textbf{x} \in \textbf{D}}{\operatorname{minimize}}& &f(\textbf{x}) \\
	\operatorname{subject\;to:}
	&&g_i(\textbf{x}) \leq 0, \quad i = 1, 2, \dots, m\\
	&&h_j(\textbf{x}) = 0, \quad j = 1, 2, \dots, p
\end{eqnarray*}
where; 1) $\mathbf{x} \in \mathbf{D} \subseteq \mathbb{R}^n$ is a candidate solution to the problem, being $n$ the problem dimensionality, and $\mathbf{D}$ the search space, typically defined in terms of bounding box constraints $lb_k \leq x_k \leq ub_k ~ \forall k \in \{1, 2, \dots, n\}$, where $lb_k$ and $ub_k$ are the lower and upper bound, respectively, for each k-th variable; 2) $f(\textbf{x}): \mathbb{R}^n \to \mathbb{R}$ is the objective function; 3) $g_i(\textbf{x})$ and $h_j(\textbf{x})$ (both defined as: $\mathbb{R}^n \to \mathbb{R}$) are respectively, inequality and equality constraints. %with $m,p \geq 0$.

In the past three decades, a large number of computational techniques has been proposed to solve efficiently this class of problems, among which Evolutionary Algorithms (EAs) \cite{Michalewicz96EAs} have shown a great potential due to their general applicability and effectiveness. So far, most of the research in the field has focused on how to improve the \emph{feasible} results obtained by EAs, for instance developing \emph{ad hoc} evolutionary operators, specific constraint repair mechanisms, or constraint handling techniques (CHTs). However, in various real-world applications it could be desirable, or at least acceptable, to consider also \emph{infeasible} solutions. This could be obtained for instance by defining different \emph{tolerance levels} for each constraint, so to reason on the effect of relaxing a certain constraint (and, if so, how much to do that) in order to obtain an improvement on the objective function, and therefore find different trade-offs in terms of constraint violations vs. objective. Despite these application needs, to date little research effort has been put on how to allow EAs to identify, rather than a single optimal solution, a \emph{diverse} set of solutions characterized by different trade-offs of this kind. In this sense, the most notable exceptions that explicitly addressed this problem --although with contrasting results-- have focused on multi-objective approaches, where the constraint violations were considered as additional objectives to be minimized \cite{surry1997comoga,Zhou2003,Venkatraman2005,hernandez2004,deb2010hybrid}, or surrogate methods \cite{Bagheri2018}.

In this paper, our goal is to evaluate the applicability of the Multi\hyph{}dimensional Archive of Phenotypic Elites (MAP-Elites) \cite{cully2015robots,Mouret2015MAPElites}, an EA recently introduced in the literature in the context of robotic tasks, for tackling these problems, specifically to provide trade-off solutions in constrained optimization. Differently from conventional EAs, MAP-Elites conducts the search by mapping the highest-performing solutions found during the search (elites) into another multi-dimensional \emph{discretized} space, defined by problem-specific features (the latter space is separate from the original search space, and typically of a lower dimensionality). These features are uncorrelated to the actual objective function, and describe some domain-specific properties of the candidate solutions. By means of this mapping, the algorithm ``illuminates'' the search space by showing the potential value of each area of the feature space, and the corresponding trade-off between the objective and the features of interest.

The functioning of MAP-Elites is simple and intuitive. First, the multi-dimensional feature space is discretized into a multi\hyph{}dimensional grid, where each bin (i.e., a cell in the grid, which is in general a hyper-rectangle) represents a different ``niche''. Then, an EA-like search is performed by means of selection and variation (mutation and crossover), but instead of keeping a population of solutions which may or may not be diverse, MAP-Elites explicitly maintains diversity by keeping in each niche one elite, which identifies the best solution characterized by the corresponding feature values. At the end of the optimization procedure, a full \emph{map} of possible solutions is provided (rather than a single optimal solution, as in conventional single-objective EAs), each characterized by different features. This map is shown in the form of a multi-dimensional heatmap, which allows for an easy visual inspection of how the objective function changes across the feature space.

In order to apply MAP-Elites to constrained optimization, the main idea we propose here is to define the feature space based on a discretization of the constraint violations. It is worth noting that in practical applications the discretization has a concrete, domain-dependent meaning: it can be seen as a set of tolerance levels, which as we mentioned can be different for each constraint. With this approach, we are able to produce a visual representation of the objective values in the feature space (in this case, space of constraint violations), thus uncovering possible correlations between the constraints and the objective. Thanks to this visualization, it is indeed easy to understand ``where'', with respect to the constraints boundaries, the best solutions lie. It also is easy to inspect the best overall solution, and check if the algorithm was able to produce particularly interesting solutions violating some of the constraints. As said, this insight can be helpful in cases where the violation of some constraints (within a certain tolerance level) can be an acceptable trade-off for a better overall performance.
%Each bin in the heatmap represents an amount of error on the constraint.
%When dealing with more than two dimensions, the higher dimensions are nested inside the bins of the first two dimensions, making it possible to have an immediate visual feedback over the exploration space.

%to summarize
As we will see in detail in the paper, despite its simplicity the proposed approach has various advantages: 1) it can be easily adapted/extended to include custom evolutionary operators; 2) it does not necessarily need explicit CHTs, but it can also include them; 3) it implicitly preserves diversity; 4) it allows the user to easily define custom tolerance levels, different for each constraint; 5) it ``illuminates'' the search space as it provides additional information on the correlation between constraints and objective, which might be of interest in practical applications; 6) it facilitates the interpretation of results through an intuitive visualization.

The rest of the paper is structured as follows. In the next Section, we will briefly summarize the most recent works on MAP-Elites and constrained optimization. Then, Section \ref{sec:algorithm} describes the basic MAP-Elites algorithm and how it can be applied to constrained optimization. In Section \ref{sec:experimental_setup}, we describe the experimental setup (benchmark and algorithmic settings), followed by the analysis of the numerical results, reported in Section \ref{sec:results}. Finally, Section \ref{sec:conclusions} concludes this work and suggests possible future developments.

\section{Related work}\label{sec:rw}
The study of MAP-Elites and, more in general, EAs explicitly driven by \emph{novelty} \cite{lehman2008exploiting,Mouret2011Multiobjectivization,shahrzad2018enhanced} or \emph{diversity} \cite{pugh2016quality,Cully2018QD}, rather than the objective alone, is a relatively new area of research in the Evolutionary Computation community. Among these algorithms, MAP-Elites \cite{cully2015robots,Mouret2015MAPElites} has attracted quite some attention in the field, due to its simplicity and general applicability. Since its introduction in 2015, MAP-Elites has been mostly used as a means to identify \emph{repertoires} of different agent behaviors e.g. in evolutionary robotics setups. Various examples of applications to maze navigation, legged robot gait optimization, and anthropomorphic robot trajectory optimization can be found in \cite{Auerbach2016GIQ, cully2015robots,Mouret2015MAPElites,Vassiliades2017,Vassiliades2018,Vassiliades2018CVT,samuelsen2018multi,Cully2018QD}. More recently, MAP-Elites has been applied also to Workforce Scheduling and Routing Problem (WSRP) \cite{Urquhart2018WSRP} and Genetic Programming \cite{dolson2018exploring}. To the best of our knowledge, no prior work exists on the explicit use of MAP-Elites for solving constrained optimization problems. %Vassiliades2017CMO

Evolutionary constrained optimization is, on the other hand, a much more mature area of research: hundreds of papers have shown in the past three decades various algorithmic solutions and real-world problems where EAs were successfully applied to constrained optimization. Summarizing all the recent advances in this area would be impossible, and is obviously outside the scope of this paper. A thorough survey of the literature is performed, for instance, in \cite{Maesani2016mVIE}, to which we refer the interested reader for a comprehensive analysis of the state-of-the-art updated to 2016. Another interesting study, published at the end of 2018 by Hellwig and Beyer \cite{hellwig2019benchmarking}, covers all the aspects related to benchmarking EAs for constrained optimization, including a thorough analysis of the most important benchmark suites available in the literature. Among these, the CEC 2010 benchmark \cite{mallipeddi2010problem} has attracted in the past few years a large body of works that showed how to solve its functions efficiently, and is often used nowadays for benchmarking new algorithms. Currently, the state-of-the-art results on this benchmark have been obtained by $\varepsilon$DEag, an $\varepsilon$ constrained Differential Evolution algorithm with an archive and gradient-based mutation proposed by Takahama and Sakai \cite{Takahama10eDEag}, and ECHT-DE, another variant of Differential Evolution that includes an ensemble of four constraint handling techniques, proposed by Mallipeddi and Suganthan \cite{mallipeddi2010differential}. These two works are also good examples of two of the most successful recent trends in the field, that are the use of gradient-based information (if available, or at least approximable), and the combination of multiple CHTs into a single evolutionary algorithm. 

% ------------------------------------------------------------

\section{Methodology}\label{sec:algorithm}
The basic version of MAP-Elites, as introduced in \cite{cully2015robots,Mouret2015MAPElites}, is shown in \cref{code:mapelites}. In the pseudo-code, $\textbf{x}$ and $\textbf{x}'$ are candidate solutions (i.e., $n$-dimensional vectors defined in the search space $\mathbf{D}$); $\textbf{b}'$ is a \emph{feature descriptor}, that is a location in a user-defined \emph{discretized} feature space, corresponding to the candidate solution $\textbf{x}'$, (i.e., an $N$-dimensional vector of user-defined features that characterize $\textbf{x}'$, typically with $N<n$); $p'$ is the performance of the candidate solution $\textbf{x}'$ (i.e., the scalar value returned by the objective function $f(\textbf{x}')$; the function itself is assumed to be a black-box, that is its mathematical formulation, if any, is unknown to the algorithm); $\mathcal{P}$ is a <feature descriptor, performance> map (i.e. an associative table that stores the best performance associated to each feature descriptor encountered by the algorithm); $\mathcal{X}$ is a <feature descriptor, solution> map (i.e. an associative table that stores the best solution associated to each feature descriptor encountered by the algorithm); $\mathcal{P}(\textbf{b}')$ is the best performance associated to the feature descriptor $\textbf{b}'$ (it can be empty); $\mathcal{X}(\textbf{b}')$ is the best solution associated to the feature descriptor $\textbf{b}'$ (it can be empty).
\begin{algorithm}
    \begin{algorithmic}
            %\Procedure{MAP-Elites}{}
            \State $\mathcal{P} \gets \emptyset, \mathcal{X} \gets \emptyset$
            \For {$iter = 1 \to I$}
                \If{$iter < G$}
                    \State $\textbf{x}' \gets \textrm{random\_solution}()$
                \Else
                    \State $\textbf{x} \gets \textrm{random\_selection}(\mathcal{X})$
                    \State $\textbf{x}' \gets \textrm{random\_variation}(\textbf{x})$
                \EndIf
                \State $\textbf{b}' \gets \textrm{feature\_descriptor}(\textbf{x}')$
                \State $p' \gets \textrm{performance}(\textbf{x}')$
                \If {$\mathcal{P}(\textbf{b}') = \emptyset \lor \mathcal{P}(\textbf{b}') > p'$}
                    \State $\mathcal{P}(\textbf{b}') \gets p'$
                    \State $\mathcal{X}(\textbf{b}') \gets \textbf{x}'$
                \EndIf
            \EndFor
            \State 
            \Return $\mathcal{P}$ and $\mathcal{X}$
            %\EndProcedure
    \end{algorithmic}
    \caption{MAP-Elites algorithm, taken from \cite{Mouret2015MAPElites}}
    \label{code:mapelites}
\end{algorithm}

Following the pseudo-code, the algorithm first creates the two maps $\mathcal{P}$ and $\mathcal{X}$, which are initially empty. Then, a loop of $I$ iterations (i.e., function evaluations) is executed. For each of the first $G$ iterations, $G$ solutions are randomly sampled in the search space $\mathbf{D}$, which are used for initializing the two maps $\mathcal{P}$ and $\mathcal{X}$. Then, starting from the iteration $G+1$, a solution $\textbf{x}$ is randomly selected from the current map $\mathcal{X}$, and a randomly modified copy of it, $\textbf{x'}$, is generated. The feature descriptor $\textbf{b}'$ and performance $p'$ associated to this new, perturbed solution are then evaluated. At this point, the two maps $\mathcal{P}$ and $\mathcal{X}$ are updated: if the performance associated to $\textbf{b}'$, $\mathcal{P}(\textbf{b}')$, is empty (which can happen if this is the first time that the algorithm generates a solution with that feature descriptor), or if it contains a value that is worse than the performance $p'$ of the newly generated solution (in \cref{code:mapelites}, we assume a minimization problem, therefore we check the condition $\mathcal{P}(\textbf{b}') > p'$), the new solution $\textbf{x'}$ and its performance $p'$ are assigned to the elements of the maps corresponding to its feature descriptor $\textbf{b'}$, namely $\mathcal{P}(\textbf{b}')$ and $\mathcal{X}(\textbf{b}')$. Once the loop terminates, the algorithm returns the two maps $\mathcal{P}$ and $\mathcal{X}$, which can be later analyzed for further inspection and post-processing.

It can be immediately noted how simple the algorithm is. With reference to the pseudo-code, in order to apply MAP-Elites to a specific problem the following methods must be defined:
\begin{itemize}
    \item \textrm{random\_solution()}: returns a randomly generated solution;
    \item $\textrm{random\_selection($\mathcal{X}$)}$: randomly selects a solution from $\mathcal{X}$;
    \item $\textrm{random\_variation(\textbf{x})}$: returns a modified copy of $\textbf{x}$;
    \item $\textrm{feature\_descriptor}(\textbf{x})$: maps a candidate solution $\textbf{x}$ to its representation in the feature space, $\textbf{b}$;
    \item $\textrm{performance}(\textbf{x})$: evaluates the objective function corresponding to the candidate solution $\textbf{x}$.
\end{itemize}
The first three methods are rather standard, i.e., they can be based on general-purpose operators typically used in EAs. However, it is possible to customize them according to the specific need. For instance, the basic version MAP-Elites randomly selects at each iteration one solution, and applies only Gaussian mutation operator; on the other hand, the algorithm can be easily configured to use a different selection mechanism (e.g. an informed operator that introduces some selection pressure/bias) or select multiple solutions at each iteration so to apply a recombination operator (crossover) or some other search mechanism such as a local search. We will see in the next Section the details of three different algorithm configurations that we have used in our experimentation.

As for what concerns the last two methods, $\textrm{feature\_descriptor}(\textbf{x})$ and $\textrm{performance}(\textbf{x})$, these are obviously problem-dependent: the first one, being dependent on how the user defines the features of interest and the corresponding feature space; the latter, being dependent on the specific objective function at hand.

The application of MAP-Elites to constrained optimization is then quite straightforward: here, we map each constraint of a constrained optimization problem to a different feature in the feature space explored by MAP-Elites, such that each candidate solution is associated to a feature descriptor that is basically a vector of constraint violations. In this specific case then, the user does not necessarily have to define any additional feature, but the features themselves are already part of the problem definition. Leaving aside the algorithmic details (selection and variation) and parameters (the only two parameters of the algorithm are the total and initial number of iterations, respectively $I$ and $G$, which can be easily set by the user based on computing resources and/or time constraints), the only input required from the user is the discretization of the features (constraint violations) space. 

An intuitive way of discretizing this space is to define, for each constraint, a certain number of \emph{tolerance levels}, i.e. amounts of constraint violation used as discretization steps. These can be easily expressed in absolute terms (based on the values of $g_i(\textbf{x})$ and $h_j(\textbf{x})$ in case of violations), or normalized w.r.t. known minimum and maximum violations. A simple example of discretization steps is $\{0, \varepsilon, 2\varepsilon, \dots\}$, where $\varepsilon$ is a user-defined parameter. However, as we will show in the next Section, also non-linear discretization is possible. In general, the discretization strategy should be based on domain knowledge and defined in such a way that solutions whose violations are equivalent, from a practical point of view, are grouped in the same bin. This would allow to ``illuminate'' the relation between objective function and constraint violations in a significant, meaningful way. Finally, we must note that while in general a different set of \emph{tolerance levels} can be defined for each constraint (especially if these are expressed in absolute terms), if all constraints have the same codomain (or, if they are normalized), the same tolerance levels can be used for all of them.

%\item \texttt{map\_x\_to\_b} (corresponding to \texttt{feature\_descriptor(x')} in the pseudo-code): maps a solution $x$ to its representation in the feature space $b_x$.
%\item \texttt{performance\_measure} (corresponding to the \texttt{performance(x')} function): evaluation of the objective function against a solution.
%\item \texttt{generate\_random\_solution} (corresponding to \texttt{random\_solution()} function): must return a randomly generated solution
%\item \texttt{generate\_feature\_dimensions}: must return a vector of \texttt{FeatureDimension} objects, which describe the features dimensions and their discretization. In our setting, we will have one object per constraint implementing the specific function and the mapping of the constraint error to a specific bin in the feature space.

% ------------------------------------------------------------

\section{Experimental Setup}\label{sec:experimental_setup}
We evaluated the performance of the proposed approach on the benchmark functions defined for the CEC 2010 Competition on Constrained Real-Parameter Optimization \cite{mallipeddi2010problem}. This benchmark presents 18 problems with different landscape characteristics, subject to a varying number (up to four) of equality and/or inequality constraints. To assess the scalability of MAP-Elites, we tested these problems in 10 and 30 dimensions. All the details of the experimental setup were set according to the CEC indications \cite{mallipeddi2010problem}. In short, the experiments were run with the following parameters:
%, in order to have a fair comparison
\begin{itemize}
    \item Number of benchmark problems: 18 \{C1, C2, $\dots$, C18\}
    \item Number of dimensions, for each problem: 10D and 30D
    \item Number of runs, for each problem/dimensionality: 25
    \item Number of function evaluations (NFEs), per run: $I=2.0e5$ for 10D, $I=6.0e5$ for 30D
    \item Number of function evaluations (NFEs) for map initialization, per run: $G=2000$ 
    \item Discretization steps for every feature: $\{0, 0.01$, $0.0001, 1\}$. This means that each feature is discretized into 5 bins, namely: $\{0\}$, $(0,0.0001]$, $(0.0001,0.01]$, $(0.01,1.0]$ and $(1.0,\inf)$.
\end{itemize}
This last aspect deserves more attention, since as we have seen in the previous Section this is what allows the application of MAP-Elites to constrained optimization. Here, we have defined as discretization steps the three tolerance levels defined in \cite{mallipeddi2010problem}, i.e. 0.0001, 0.01, and 1.0 (see the next Section), in addition to an explicit step corresponding to zero-tolerance (corresponding to solutions with $g_i(\textbf{x}) \leq 0$ and $h_j(\textbf{x}) = 0$, respectively for inequality and equality constraints).

The features used in MAP-Elites follow the same order as they appear in the corresponding problem definition, with inequality constraints considered before equality constraints. Since the problems contained in the the CEC 2010 benchmark have a variable number of equality/inequality constraints, we define a variable-sized feature space, where for each problem there are as many MAP-Elites features as constraints. For visualization purposes, we represent the map $\mathcal{P}(\textbf{b}')$ obtained in each run of MAP-Elites in the form of a multi-dimensional heatmap, as explained in \cite{cully2015robots,Mouret2015MAPElites}. The color represents the objective value corresponding to the solution contained in each bin. The 1st axis (abscissa) corresponds to the 1st constraint violation, the 2nd axis (ordinate) corresponds to the 2nd constraint violation, and so on for the 3rd and 4th constraints. Feature dimensions are ``nested'' such that the feature space is first discretized along the 1st and 2nd axes (so to obtain a 2D grid of bins), while the following features (if any) are represented by an ``inner'' (1D or 2D) discretization inside each bin in the ``outer'' grid. Obviously, this visualization procedure can be easily extended to handle more than four constraints, although the visual interpretability of the results tends to decrease with the number of features shown in the heatmap.

\begin{figure*}[ht]
\centering
    \includegraphics[width=0.69\columnwidth,trim=2cm 2cm 3cm 2.9cm,clip=true]{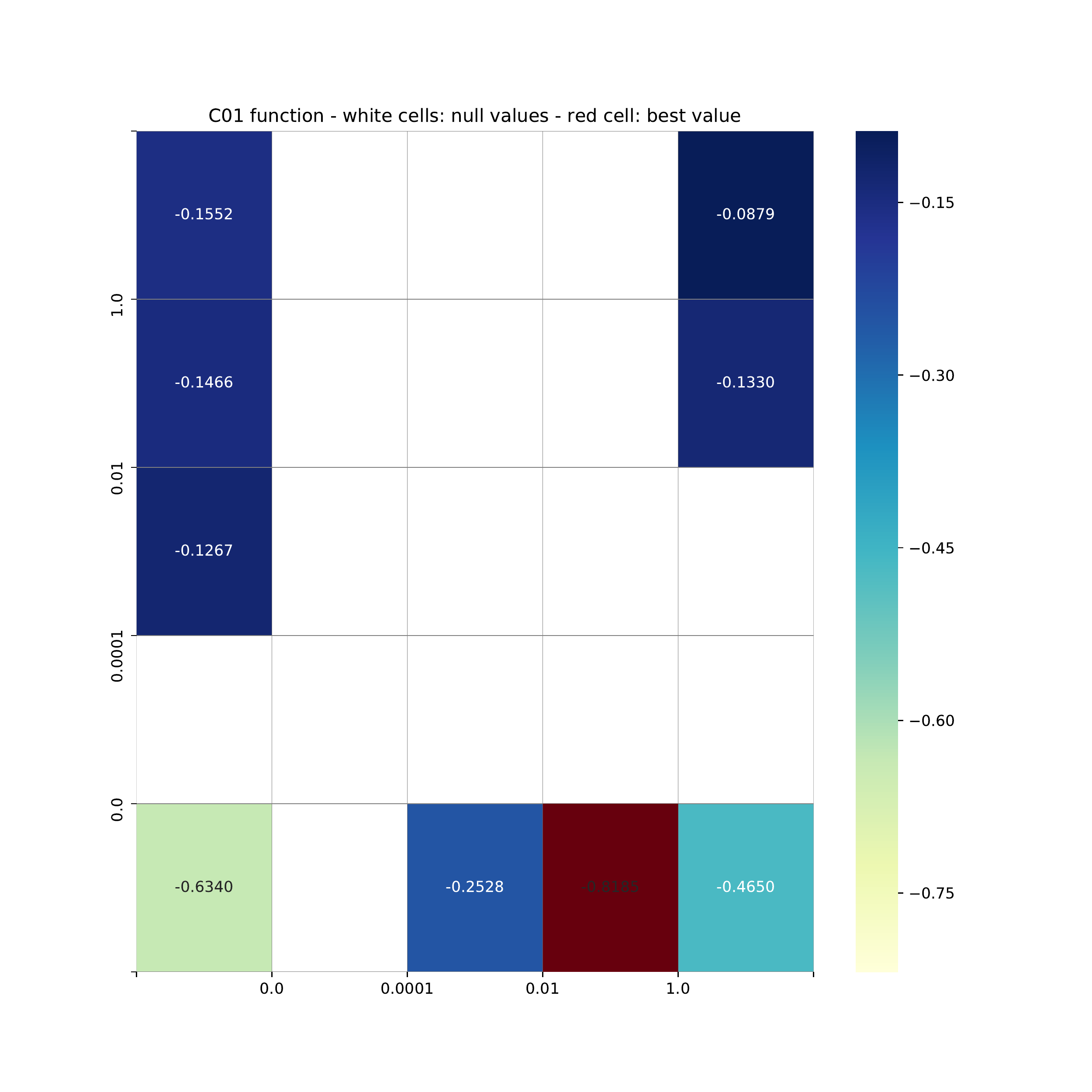}
    \includegraphics[width=0.69\columnwidth,trim=2cm 2cm 3cm 2.9cm,clip=true]{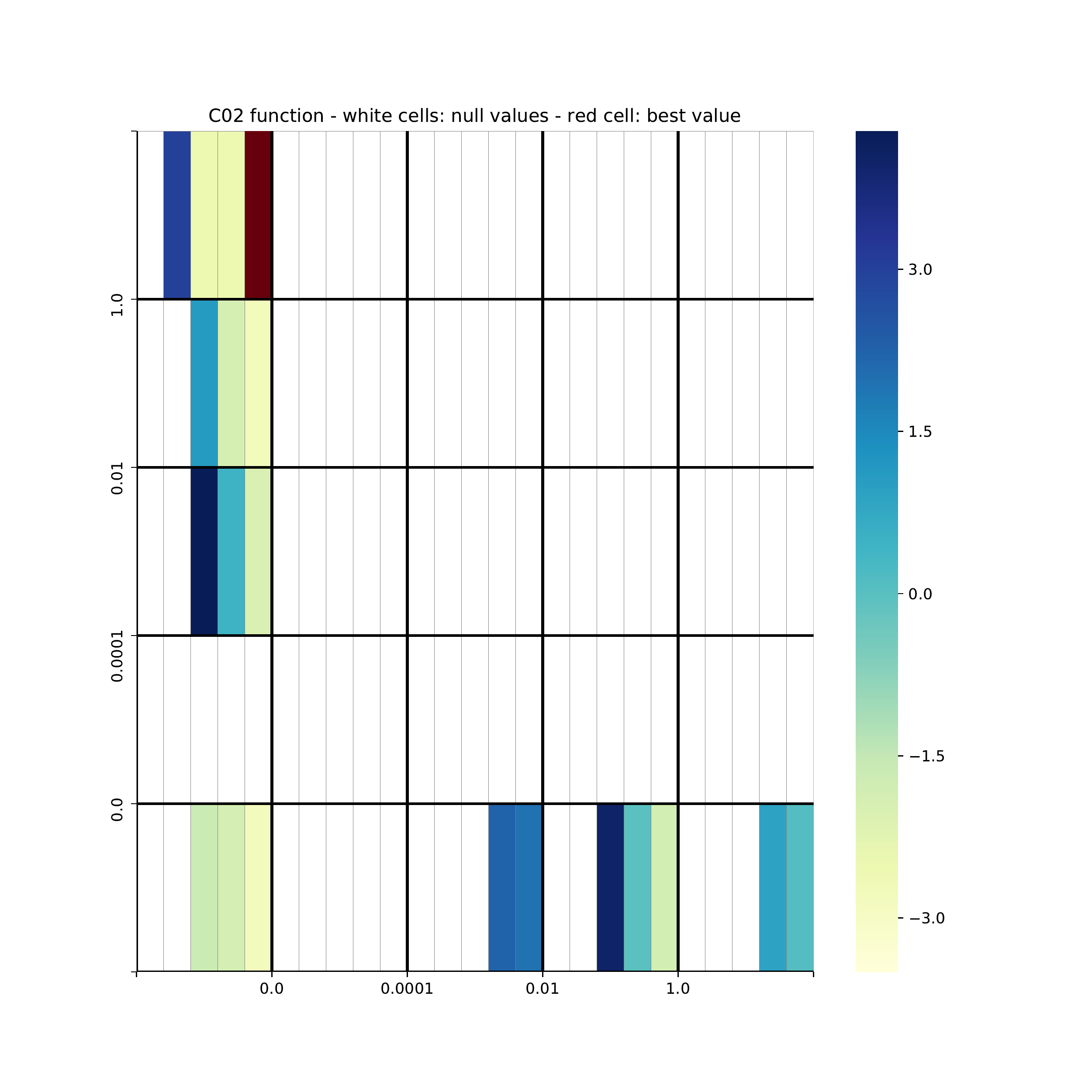}
    \includegraphics[width=0.69\columnwidth,trim=2cm 2cm 3cm 2.9cm,clip=true]{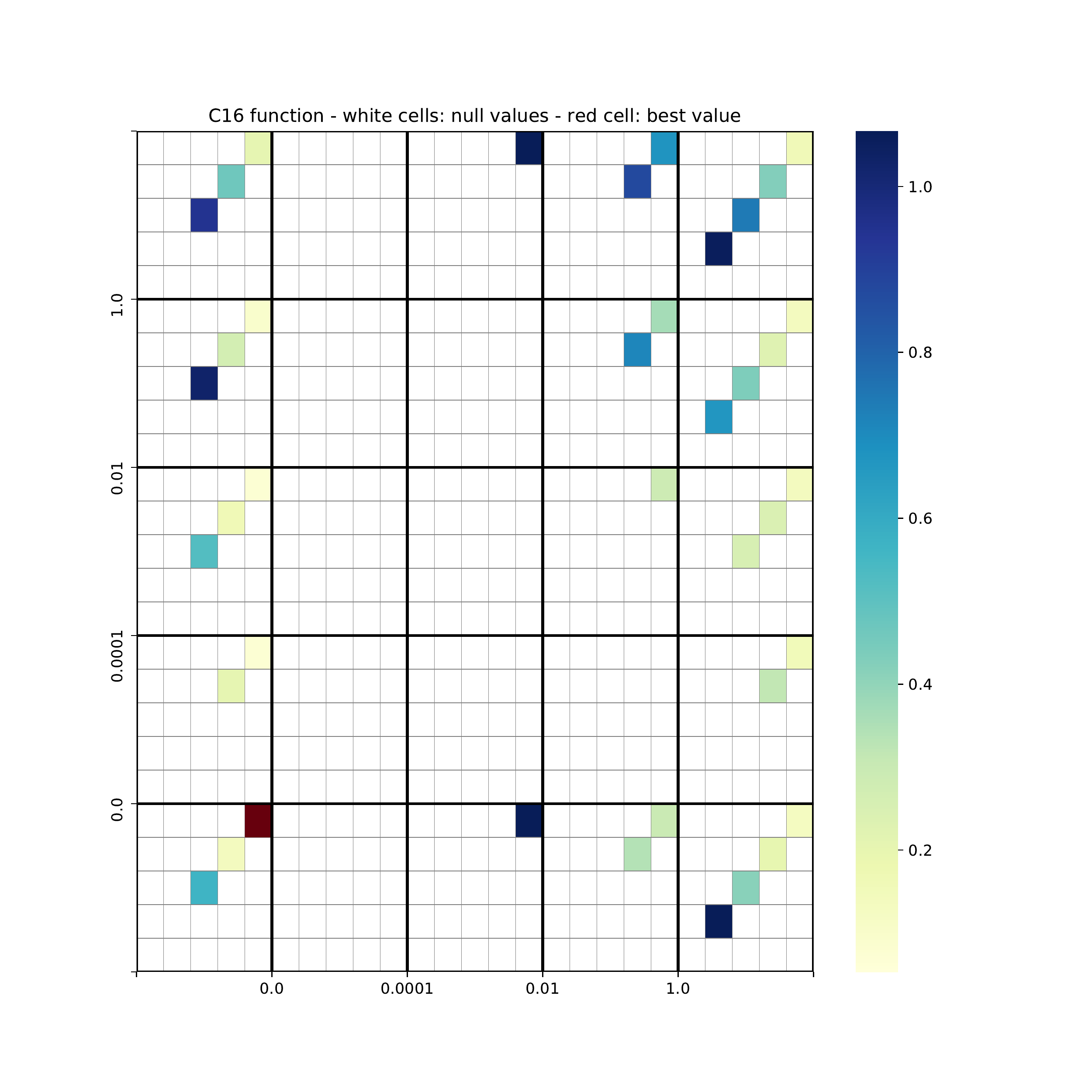}
    \caption{Final heatmaps found in a single run of MAP-Elites on C1, Configuration 1 (left), C2, Configuration 1 (center), C16, Configuration 3 (right) in 10D. The three benchmark functions are characterized, respectively, by 2I, 2I-1E, and 2I-2E, where `I' and `E' stand for inequality and equality constraints respectively. In each heatmap, the color of each bin is proportionate to the objective value corresponding to the solution present in it (assuming minimization, the lower the better), while the red bin indicates the solution with the best objective value (regardless its feasibility). It can be observed that the maps easily allow to ``illuminate'' the search space of each problem, identifying various trade-off solutions in terms of objective value vs. constraint violations, such as solutions with a high performance but with some violated constraints. Note that in case of 3 or 4 constraints the discretization along the first two (outer) dimensions of the heatmap is indicated by a thicker black line, while the discretization along the ``nested'' (inner) dimensions is indicated by a thinner black line.}
    \label{fig:heatmaps}
\end{figure*}

It should be noted that, according to the CEC 2010 benchmark definition \cite{mallipeddi2010problem}, a solution $\textbf{x}$ is considered feasible iff $g_i(\textbf{x}) \leq 0 ~\forall i \in \{1, 2, \dots, m\}$ and $|h_j(\textbf{x})| - \epsilon \leq 0 ~\forall j \in \{1, 2, \dots, p\}$, where $\epsilon$ is the equality constraint tolerance, set to 0.0001. Otherwise, the solution is considered infeasible. From what we have just discussed, it follows then that feasible solutions in the sense of the CEC 2010 definition can be found: 1) for what concerns inequality constraints, in the first bin ($\{0\}$) along each feature dimension; 2) for what concerns equality constraints, in the first two bins ($\{0\}$, $(0,0.0001]$). In plain terms, this means that we can easily identify feasible solutions found by MAP-Elites by simply looking at the lower-left corner of the heatmap, while solutions with increasing constraint violations are found scanning the heatmap (and each inner bin in case of more than two constraints) towards the upper-right side. Some examples of heatmaps obtained by MAP-Elites are shown in \cref{fig:heatmaps}.

As for the evolutionary operators (selection and variation, as shown in \cref{code:mapelites}), we defined three different algorithmic settings. In all cases, selection is performed according to a uniform distribution over the current map. Variation is instead applied according to the following configurations:
\begin{itemize}
\item Configuration 1: mutation ($\sigma=0.1$), without crossover
\item Configuration 2: mutation ($\sigma=0.5$), without crossover
\item Configuration 3: mutation ($\sigma=0.1$), with crossover
\end{itemize}
In all three cases, mutation is implemented by applying to the selected solution (with probability $0.5$ for each variable) a Gaussian mutation with $\mu=0$ and the given value of $\sigma$. Boundary constraints are handled according to a toroidal mechanism: given a decision variable $x$ constrained to the interval $[a,b]$, if the corresponding mutated variable $x'$ exceeds the upper bound $b$ (i.e., $x'=b+\zeta$), $x'$ is transformed into $x'=a+\zeta$, $\zeta>0$. Similarly, if $x'=a-\zeta$, $x'$ is transformed into $x'=b-\zeta$, $\zeta>0$.

In Configuration 3, at each iteration two solutions are randomly selected from the current map, after which uniform crossover (with probability $0.5$ for each variable) is applied by swapping the corresponding variables from the two parents. Then, the first of the two offspring generated by crossover undergoes Gaussian mutation, as in Configurations 1 and 2, and is evaluated in terms of feature descriptor and performance, as shown in \cref{code:mapelites}.

The entire experimental setup was implemented in Python 3\footnote{
\ifdefined\ANONYMOUS
    Link omitted for double-blind review.
\else
    Code available at: \url{https://github.com/StefanoFioravanzo/MAP-Elites}.
\fi
}, and the experimentation was performed on a Ubuntu 18.10 workstation, with a CPU Intel Core i9-7940X @3.10GHz and 64GB DDR4.

\section{Numerical results}\label{sec:results}
We present here the results obtained on the experimental setup described in the previous Section. In Tables \ref{tab:10D_standard_cec}-\ref{tab:30D_crossover_cec}, we report the results for all the CEC 2010 functions in 10 and 30 dimensions, for the three algorithm settings described above\footnote{ The complete set of numerical results and the final heatmaps for each problem and dimensionality are available as Supplementary Material at: \url{https://bit.ly/2WwgeTU}.}.
In the tables, we report the results as suggested in \cite{mallipeddi2010problem}, where for each function we show:
\begin{enumerate}
    \item The objective value corresponding to the best, worst, and median solution\footnote{ The final solutions are sorted according to these three criteria: 1) feasible solutions are sorted in front of infeasible solutions; 2) feasible solutions are sorted according to their objective value;
    3) infeasible solutions are sorted according to their mean value of constraint violation, calculated as in Eq. (\ref{eq:v}).} (over 25 runs) obtained at the end of the computational budget; next to each objective value, we show in parenthesis the no. of violated constraints corresponding to each of these three solutions.
    \item The number of violated constraints at the median solution, $c=(c_1, c_2, c_3)$ (where each element $c_i, i=1,2,3$ represents the number of violations higher than three tolerance levels set to 1, 0.01, and 0.0001, respectively), and the corresponding mean violation $\bar{v}$, calculated as:
    \begin{equation}\label{eq:v}
        \bar{v}=\frac{\sum_{i=1}^{m}{G_i(x)}+\sum_{j=1}^{p}{H_j(x)}}{m+p}
    \end{equation}
    where $G_i(x)=g_i(x)$ if $g_i(x)>0$ (otherwise zero), and $H_j(x)=|h_j(x)|$ if $|h_j(x)|-\epsilon>0$ (otherwise zero), being $\epsilon$ is the equality constraint tolerance (as seen earlier, 0.0001).
    \item The average objective value (over 25 runs) of the final solutions obtained at the end of the budget, and its std. dev.
    \item The Feasibility Rate (FR), that is, for each function, the ratio between the number of runs during which at least one feasible solution was found within the budget, and the total number of runs (in our case, 25).
\end{enumerate}

From the tables, we can observe that in all three configurations, MAP-Elites solves with 100\% FR C1, C7, C8, C14, C15 in 10D, i.e. all the functions with inequality constraints only (except C13, that is however the only function with inequality constraints only whose volume of the feasible region is approximately zero); in 30D, it also finds feasible solutions on C18 in 100\% of the runs in all configurations (92\% in 10D for Configuration 3). The peculiarity of this function is that, despite it has one equality constraint, the volume of its feasible region is non-zero. The only other functions on which a non-zero FR is obtained, although not in all configurations and dimensionalities, are: C2, C9, C10, C16, C17. Except C10 that has 1 "rotated" equality constraint, all other functions have only separable constraints, which could explain why in some cases even by Gaussian mutation only (which acts independently on each variable) it is possible to reach the feasible region. Overall, among the 3 configurations Configuration 3 has the highest FR across all the tests, while it results that an excessively high value of $\sigma$ in Gaussian mutations (as in Configuration 2) is detrimental.

From these observations, we can conclude that the basic MAP-Elites algorithm we have used in our experimentation is not able to solve efficiently either problems with non-separable equality constraints, or with an approximately zero-volume feasibile region. This is not surprising though, as the algorithm is only based on simple genetic operators (Gaussian mutations and uniform crossover in our case) that do not use any information about the constraints. In contrast, the two best-performing algorithms on the CEC 2010 benchmark, $\varepsilon$DEag \cite{Takahama10eDEag} and ECHT-DE \cite{mallipeddi2010differential}, encapsulate highly efficient CHTs (and, in the case of $\varepsilon$DEag, gradient information about constraints) that allow them to reach an 100\% FR on all functions in 10D and 30D (except, respectively for the two algorithms, C12 in 30D, and C11-C12 in both 10D and 30D), as reported in the original papers.

This comparison encourages though the idea to explore in the future the possibility to include into the MAP-Elites scheme at least one dedicated technique for better handling equality constraints, such as the $\varepsilon$ constrained method, initially introduced in \cite{takahama2005constrained} and since then used in most of the state-of-the-art algorithms for constrained optimization. Notably, the strength of this method is that it guides the search by allowing $\varepsilon$ level comparisons with a progressively shrinking relaxation (defined by the $\varepsilon$ parameter) of the constraint boundaries.

Considering the objective values, similar considerations can be drawn: limiting the analysis on the functions with 100\% FR, it results that MAP-Elites is less efficient at finding optimal values than $\varepsilon$DEag \cite{Takahama10eDEag} (the best algorithm on the CEC 2010 benchmark). In all cases MAP-Elites is several orders of magnitude worse than $\varepsilon$DEag, except C1 in 10D where instead the configuration with crossover finds a better optimal value\footnote{ We refer the interested reader to the Supplementary Material online, where we present the results after $I=2.0e5$ and $I=6.0e5$ NFEs, respectively for 10D and 30D, obtained by $\varepsilon$DEag (taken from \cite{Takahama10eDEag}). We also show a detailed report of the MAP-Elites results focused on a fitness-based rank, rather than the rank based on the sorting criteria described in the text. These results are omitted here due to space limitations.}. Once again, this conclusion is not surprising and is also in line with what was observed by Runarsson and Yao \cite{Runarsson2005}, who identified the reason for the sometimes poor results obtained by multi-objective approaches (such as \cite{surry1997comoga,Zhou2003,Venkatraman2005,hernandez2004}): in fact, when applied to constrained optimization, the Pareto ranking leads to a ``bias-free'' search that is not able to properly guide the search towards (and within) the feasible region. In other words, allowing the search to spend too many evaluations in the infeasible region makes it harder to find feasible solutions, but also to find feasible solutions with optimal values of the objective function. This might be the case also of MAP-Elites, where some form of bias (such as the $\varepsilon$ constrained method) might be needed.

% ------------------------------------------------------------

\section{Conclusions}\label{sec:conclusions}
In this paper we have explored the use of MAP-Elites for solving constrained continuous optimization problems. In the proposed approach, each feature in the feature space explored by MAP-Elites corresponds, quite straightforwardly, to the violation of each constraint, discretized according to user-defined steps (tolerance levels). In this way, the algorithm allows to ``illuminate'' the search space and thus uncover possible correlations between the constraints and the objective. The visualization of MAP-Elites also gives users the possibility to focus on different solutions characterized by different values of constraint violations. We have tested this approach on a large number of benchmark problems in 10 and 30 dimensions, characterized by up to four equality/inequality constraints. Our numerical results showed that while MAP-Elites obtains results that are not particularly competitive with the state-of-the-art on all problems (especially those with equality constraints), it is still able to provide new valuable, easy-to-understand information that can be of great interest for practitioners. Additionally, the algorithm can be easily implemented and applied without any specific tuning to various real-world problems, for instance in engineering design, where different tolerance levels can be defined depending on the specific constraints.

Since our goal was to evaluate the performance of the basic MAP-Elites on constrained optimization, the proposed approach is purposely quite simplistic, but clearly it can be extended in various ways. First of all, the basic MAP-Elites algorithm we used in this work (as shown in \cref{code:mapelites} can be replaced with some more advanced variants recently proposed in the literature. In particular, the version of MAP-Elites based on centroidal Voronoi tessellation (CVT-MAP-Elites) \cite{Vassiliades2018CVT} can be used instead of the basic one in order to scale the algorithm to a larger number of constraints. In order to better handle the unbounded feature spaces (thus avoiding the need for an explicit ``upper'' bin, $(1.0,\inf)$ in our case), the ``expansive'' MAP-Elites variants introduced in \cite{Vassiliades2017} can be employed instead, which are able to expand their bounds (in the feature space) based on the discovered solutions. Other possibilities will be to use the ``directional variation'' operator introduced in \cite{Vassiliades2018}, that exploits inter-species (or inter-elites) correlations to accelerate the search, add, most of all, specific constraint handling techniques (especially for handling equality constraints, which as we have seen is the main weakness of this approach) \cite{mallipeddi2010differential,gong2015adaptive}. It is also worth considering the use of surrogate models, such as in \cite{wangGlobalLocalSurrogate2018}, in order to further speed up the the search and guide it towards the feasible region, still allowing the algorithm to keep infeasible solutions as part of the map. Another improvement can be obtained by using an explicit Quality Diversity measure \cite{Auerbach2016GIQ,Pugh2015CCQ,pugh2016quality,Cully2018QD}, so to enforce at the same time a better coverage of the feature space and a further improvement in terms of optimization results. It is also possible to hybridize the basic MAP-Elites algorithm with local search techniques (such that MAP-Elites explores the feature space and local search is applied within one bin to further refine the search), or to devise memetic computing approaches based on combination of MAP-Elites and other metaheuristics, such as CMA-ES, that has been recently applied successfully also to constrained optimization \cite{deMelo2014cmaPenalty,jamal2018solving, hellwig2018matrix}.

Finally, on the application side it will be interesting to evaluate the applicability of this approach on combinatorial constrained optimization problems, which can be obtained by simply modifying the variation operators, or multi-objective constrained optimization, which can be obtained by adding a Pareto-dominance check, as recently shown in the context of robotic experiments \cite{samuelsen2018multi}.

% ------------------------------------------------------------

% NOTE: references don't count on the page limit - at most 8 pages (excluding references) 
% see https://gecco-2019.sigevo.org/index.html/Call+for+Papers

%\clearpage
\balance
\bibliographystyle{ACM-Reference-Format}
\bibliography{bib/references_new,bib/references_constr_optim}


\begin{thebibliography}{34}

%%% the Web address.

\ifx \showCODEN    \undefined \def \showCODEN     #1{\unskip}     \fi
\ifx \showDOI      \undefined \def \showDOI       #1{#1}\fi
\ifx \showISBNx    \undefined \def \showISBNx     #1{\unskip}     \fi
\ifx \showISBNxiii \undefined \def \showISBNxiii  #1{\unskip}     \fi
\ifx \showISSN     \undefined \def \showISSN      #1{\unskip}     \fi
\ifx \showLCCN     \undefined \def \showLCCN      #1{\unskip}     \fi
\ifx \shownote     \undefined \def \shownote      #1{#1}          \fi
\ifx \showarticletitle \undefined \def \showarticletitle #1{#1}   \fi
\ifx \showURL      \undefined \def \showURL       {\relax}        \fi
% invisible to TeX
\providecommand\bibfield[2]{#2}
\providecommand\bibinfo[2]{#2}
\providecommand\natexlab[1]{#1}
\providecommand\showeprint[2][]{arXiv:#2}

\bibitem[\protect\citeauthoryear{Auerbach, Iacca, and Floreano}{Auerbach
  et~al\mbox{.}}{2016}]%
        {Auerbach2016GIQ}
\bibfield{author}{\bibinfo{person}{Joshua~E. Auerbach},
  \bibinfo{person}{Giovanni Iacca}, {and} \bibinfo{person}{Dario Floreano}.}
  \bibinfo{year}{2016}\natexlab{}.
\newblock \showarticletitle{Gaining Insight into Quality Diversity}. In
  \bibinfo{booktitle}{\emph{Proceedings of the Genetic and Evolutionary
  Computation Conference (GECCO) Companion}}. \bibinfo{publisher}{ACM},
  \bibinfo{address}{New York, NY, USA}, \bibinfo{pages}{1061--1064}.
\newblock


\bibitem[\protect\citeauthoryear{Bagheri, Konen, and B\"ack}{Bagheri
  et~al\mbox{.}}{2018}]%
        {Bagheri2018}
\bibfield{author}{\bibinfo{person}{Samineh Bagheri}, \bibinfo{person}{Wolfgang
  Konen}, {and} \bibinfo{person}{Thomas B\"ack}.}
  \bibinfo{year}{2018}\natexlab{}.
\newblock \showarticletitle{How to Solve the Dilemma of Margin-Based Equality
  Handling Methods}. In \bibinfo{booktitle}{\emph{Proceedings of the Workshop
  Computational Intelligence}}, \bibfield{editor}{\bibinfo{person}{Frank
  Hoffmann}, \bibinfo{person}{Eyke H\"ullermeier}, {and} \bibinfo{person}{Ralf
  Mikut}} (Eds.). \bibinfo{publisher}{KIT Scientific Publishing},
  \bibinfo{address}{Karlsruhe}, \bibinfo{pages}{257--270}.
\newblock


\bibitem[\protect\citeauthoryear{Cully, Clune, Tarapore, and Mouret}{Cully
  et~al\mbox{.}}{2015}]%
        {cully2015robots}
\bibfield{author}{\bibinfo{person}{Antoine Cully}, \bibinfo{person}{Jeff
  Clune}, \bibinfo{person}{Danesh Tarapore}, {and}
  \bibinfo{person}{Jean-Baptiste Mouret}.} \bibinfo{year}{2015}\natexlab{}.
\newblock \showarticletitle{Robots that can adapt like animals}.
\newblock \bibinfo{journal}{\emph{Nature}} \bibinfo{volume}{521},
  \bibinfo{number}{7553} (\bibinfo{year}{2015}), \bibinfo{pages}{503}.
\newblock


\bibitem[\protect\citeauthoryear{Cully and Demiris}{Cully and Demiris}{2018}]%
        {Cully2018QD}
\bibfield{author}{\bibinfo{person}{Antoine Cully} {and}
  \bibinfo{person}{Yiannis Demiris}.} \bibinfo{year}{2018}\natexlab{}.
\newblock \showarticletitle{Quality and Diversity Optimization: A Unifying
  Modular Framework}.
\newblock \bibinfo{journal}{\emph{IEEE Transactions on Evolutionary
  Computation}} \bibinfo{volume}{22}, \bibinfo{number}{2}
  (\bibinfo{year}{2018}), \bibinfo{pages}{245--259}.
\newblock
\showISSN{1089-778X}


\bibitem[\protect\citeauthoryear{de~Melo and Iacca}{de~Melo and Iacca}{2014}]%
        {deMelo2014cmaPenalty}
\bibfield{author}{\bibinfo{person}{Vin\'icius~Veloso de Melo} {and}
  \bibinfo{person}{Giovanni Iacca}.} \bibinfo{year}{2014}\natexlab{}.
\newblock \showarticletitle{A modified Covariance Matrix Adaptation Evolution
  Strategy with adaptive penalty function and restart for constrained
  optimization}.
\newblock \bibinfo{journal}{\emph{Expert Systems with Applications}}
  \bibinfo{volume}{41}, \bibinfo{number}{16} (\bibinfo{year}{2014}),
  \bibinfo{pages}{7077--7094}.
\newblock


\bibitem[\protect\citeauthoryear{Deb and Datta}{Deb and Datta}{2010}]%
        {deb2010hybrid}
\bibfield{author}{\bibinfo{person}{Kalyanmoy Deb} {and}
  \bibinfo{person}{Rituparna Datta}.} \bibinfo{year}{2010}\natexlab{}.
\newblock \bibinfo{booktitle}{\emph{A hybrid bi-objective evolutionary-penalty
  approach for computationally fast and accurate constrained optimization}}.
\newblock \bibinfo{type}{{T}echnical {R}eport}. \bibinfo{institution}{KanGAL}.
\newblock


\bibitem[\protect\citeauthoryear{Dolson, Lalejini, and Ofria}{Dolson
  et~al\mbox{.}}{2018}]%
        {dolson2018exploring}
\bibfield{author}{\bibinfo{person}{Emily Dolson}, \bibinfo{person}{Alexander
  Lalejini}, {and} \bibinfo{person}{Charles Ofria}.}
  \bibinfo{year}{2018}\natexlab{}.
\newblock \showarticletitle{Exploring genetic programming systems with
  MAP-Elites}.
\newblock \bibinfo{journal}{\emph{PeerJ Preprints}}  \bibinfo{volume}{6}
  (\bibinfo{year}{2018}), \bibinfo{pages}{1--18}.
\newblock


\bibitem[\protect\citeauthoryear{Gong, Cai, and Liang}{Gong
  et~al\mbox{.}}{2015}]%
        {gong2015adaptive}
\bibfield{author}{\bibinfo{person}{Wenyin Gong}, \bibinfo{person}{Zhihua Cai},
  {and} \bibinfo{person}{Dingwen Liang}.} \bibinfo{year}{2015}\natexlab{}.
\newblock \showarticletitle{Adaptive ranking mutation operator based
  differential evolution for constrained optimization}.
\newblock \bibinfo{journal}{\emph{IEEE Transactions on Cybernetics}}
  \bibinfo{volume}{45}, \bibinfo{number}{4} (\bibinfo{year}{2015}),
  \bibinfo{pages}{716--727}.
\newblock


\bibitem[\protect\citeauthoryear{Hellwig and Beyer}{Hellwig and Beyer}{2018}]%
        {hellwig2018matrix}
\bibfield{author}{\bibinfo{person}{Michael Hellwig} {and}
  \bibinfo{person}{Hans-Georg Beyer}.} \bibinfo{year}{2018}\natexlab{}.
\newblock \showarticletitle{A matrix adaptation evolution strategy for
  constrained real-parameter optimization}. In
  \bibinfo{booktitle}{\emph{Congress on Evolutionary Computation (CEC)}}.
  \bibinfo{publisher}{IEEE}, \bibinfo{address}{New York},
  \bibinfo{pages}{1--8}.
\newblock


\bibitem[\protect\citeauthoryear{Hellwig and Beyer}{Hellwig and Beyer}{2019}]%
        {hellwig2019benchmarking}
\bibfield{author}{\bibinfo{person}{Michael Hellwig} {and}
  \bibinfo{person}{Hans-Georg Beyer}.} \bibinfo{year}{2019}\natexlab{}.
\newblock \showarticletitle{Benchmarking evolutionary algorithms for single
  objective real-valued constrained optimization--A critical review}.
\newblock \bibinfo{journal}{\emph{Swarm and evolutionary computation}}
  \bibinfo{volume}{44} (\bibinfo{year}{2019}), \bibinfo{pages}{927--944}.
\newblock


\bibitem[\protect\citeauthoryear{Hern{\'a}andez~Aguirre, Botello~Rionda,
  Liz{\'a}arraga~Liz{\'a}arraga, and Coello~Coello}{Hern{\'a}andez~Aguirre
  et~al\mbox{.}}{2004}]%
        {hernandez2004}
\bibfield{author}{\bibinfo{person}{Arturo Hern{\'a}andez~Aguirre},
  \bibinfo{person}{Salvador Botello~Rionda}, \bibinfo{person}{Giovanni
  Liz{\'a}arraga~Liz{\'a}arraga}, {and} \bibinfo{person}{Carlos
  Coello~Coello}.} \bibinfo{year}{2004}\natexlab{}.
\newblock \showarticletitle{IS-PAES: Multiobjective Optimization with Efficient
  Constraint Handling}. In \bibinfo{booktitle}{\emph{IUTAM Symposium on
  Evolutionary Methods in Mechanics}},
  \bibfield{editor}{\bibinfo{person}{Tadeusz Burczy{\'{n}}ski} {and}
  \bibinfo{person}{Andrzej Osyczka}} (Eds.). \bibinfo{publisher}{Springer},
  \bibinfo{address}{Dordrecht}, \bibinfo{pages}{111--120}.
\newblock


\bibitem[\protect\citeauthoryear{Jamal, Ming, and Zhengang}{Jamal
  et~al\mbox{.}}{2018}]%
        {jamal2018solving}
\bibfield{author}{\bibinfo{person}{Miftah~Bedru Jamal}, \bibinfo{person}{Fang
  Ming}, {and} \bibinfo{person}{Jiang Zhengang}.}
  \bibinfo{year}{2018}\natexlab{}.
\newblock \showarticletitle{Solving constrained optimization problems by using
  covariance matrix adaptation evolutionary strategy with constraint handling
  methods}. In \bibinfo{booktitle}{\emph{Proceedings of the International
  Conference on Innovation in Artificial Intelligence}}.
  \bibinfo{publisher}{ACM}, \bibinfo{address}{New York, NY, USA},
  \bibinfo{pages}{6--15}.
\newblock


\bibitem[\protect\citeauthoryear{Lehman and Stanley}{Lehman and
  Stanley}{2008}]%
        {lehman2008exploiting}
\bibfield{author}{\bibinfo{person}{Joel Lehman} {and}
  \bibinfo{person}{Kenneth~O. Stanley}.} \bibinfo{year}{2008}\natexlab{}.
\newblock \showarticletitle{Exploiting open-endedness to solve problems through
  the search for novelty}. In \bibinfo{booktitle}{\emph{International
  conference on Artificial Life (Alife XI)}}. \bibinfo{publisher}{MIT Press},
  \bibinfo{address}{Cambridge, MA}, \bibinfo{pages}{329--336}.
\newblock


\bibitem[\protect\citeauthoryear{Maesani, Iacca, and Floreano}{Maesani
  et~al\mbox{.}}{2016}]%
        {Maesani2016mVIE}
\bibfield{author}{\bibinfo{person}{Andrea Maesani}, \bibinfo{person}{Giovanni
  Iacca}, {and} \bibinfo{person}{Dario Floreano}.}
  \bibinfo{year}{2016}\natexlab{}.
\newblock \showarticletitle{Memetic Viability Evolution for Constrained
  Optimization}.
\newblock \bibinfo{journal}{\emph{IEEE Transactions on Evolutionary
  Computation}} \bibinfo{volume}{20}, \bibinfo{number}{1}
  (\bibinfo{year}{2016}), \bibinfo{pages}{125--144}.
\newblock
\showISSN{1089-778X}


\bibitem[\protect\citeauthoryear{Mallipeddi and Suganthan}{Mallipeddi and
  Suganthan}{2010a}]%
        {mallipeddi2010differential}
\bibfield{author}{\bibinfo{person}{Rammohan Mallipeddi} {and}
  \bibinfo{person}{Ponnuthurai~Nagaratnam Suganthan}.}
  \bibinfo{year}{2010}\natexlab{a}.
\newblock \showarticletitle{Differential evolution with ensemble of constraint
  handling techniques for solving CEC 2010 benchmark problems}. In
  \bibinfo{booktitle}{\emph{Congress on Evolutionary Computation (CEC)}}.
  \bibinfo{publisher}{IEEE}, \bibinfo{address}{New York},
  \bibinfo{pages}{1--8}.
\newblock


\bibitem[\protect\citeauthoryear{Mallipeddi and Suganthan}{Mallipeddi and
  Suganthan}{2010b}]%
        {mallipeddi2010problem}
\bibfield{author}{\bibinfo{person}{Rammohan Mallipeddi} {and}
  \bibinfo{person}{Ponnuthurai~Nagaratnam Suganthan}.}
  \bibinfo{year}{2010}\natexlab{b}.
\newblock \bibinfo{booktitle}{\emph{Problem definitions and evaluation criteria
  for the CEC 2010 competition on constrained real-parameter optimization}}.
\newblock \bibinfo{type}{{T}echnical {R}eport}. \bibinfo{institution}{NTU,
  Singapore}.
\newblock


\bibitem[\protect\citeauthoryear{Michalewicz and Schoenauer}{Michalewicz and
  Schoenauer}{1996}]%
        {Michalewicz96EAs}
\bibfield{author}{\bibinfo{person}{Zbigniew Michalewicz} {and}
  \bibinfo{person}{Marc Schoenauer}.} \bibinfo{year}{1996}\natexlab{}.
\newblock \showarticletitle{Evolutionary Algorithms for Constrained Parameter
  Optimization Problems}.
\newblock \bibinfo{journal}{\emph{Evolutionary Computation}}
  \bibinfo{volume}{4}, \bibinfo{number}{1} (\bibinfo{year}{1996}),
  \bibinfo{pages}{1--32}.
\newblock
\showISSN{1063-6560}


\bibitem[\protect\citeauthoryear{Mouret}{Mouret}{2011}]%
        {Mouret2011Multiobjectivization}
\bibfield{author}{\bibinfo{person}{Jean-Baptiste Mouret}.}
  \bibinfo{year}{2011}\natexlab{}.
\newblock \showarticletitle{Novelty-Based Multiobjectivization}. In
  \bibinfo{booktitle}{\emph{New Horizons in Evolutionary Robotics}},
  \bibfield{editor}{\bibinfo{person}{St{\'e}phane Doncieux},
  \bibinfo{person}{Nicolas Bred{\`e}che}, {and} \bibinfo{person}{Jean-Baptiste
  Mouret}} (Eds.). \bibinfo{publisher}{Springer}, \bibinfo{address}{Berlin,
  Heidelberg}, \bibinfo{pages}{139--154}.
\newblock


\bibitem[\protect\citeauthoryear{Mouret and Clune}{Mouret and Clune}{2015}]%
        {Mouret2015MAPElites}
\bibfield{author}{\bibinfo{person}{Jean-Baptiste Mouret} {and}
  \bibinfo{person}{Jeff Clune}.} \bibinfo{year}{2015}\natexlab{}.
\newblock \showarticletitle{Illuminating search spaces by mapping elites}.
\newblock \bibinfo{journal}{\emph{CoRR}}  \bibinfo{volume}{abs/1504.04909}
  (\bibinfo{year}{2015}), \bibinfo{pages}{1--15}.
\newblock
\showeprint[arxiv]{1504.04909}


\bibitem[\protect\citeauthoryear{Pugh, Soros, and Stanley}{Pugh
  et~al\mbox{.}}{2016}]%
        {pugh2016quality}
\bibfield{author}{\bibinfo{person}{Justin~K. Pugh}, \bibinfo{person}{Lisa~B.
  Soros}, {and} \bibinfo{person}{Kenneth~O. Stanley}.}
  \bibinfo{year}{2016}\natexlab{}.
\newblock \showarticletitle{Quality diversity: A new frontier for evolutionary
  computation}.
\newblock \bibinfo{journal}{\emph{Frontiers in Robotics\&AI}}
  \bibinfo{volume}{3} (\bibinfo{year}{2016}), \bibinfo{pages}{40}.
\newblock


\bibitem[\protect\citeauthoryear{Pugh, Soros, Szerlip, and Stanley}{Pugh
  et~al\mbox{.}}{2015}]%
        {Pugh2015CCQ}
\bibfield{author}{\bibinfo{person}{Justin~K. Pugh}, \bibinfo{person}{L.~B.
  Soros}, \bibinfo{person}{Paul~A. Szerlip}, {and} \bibinfo{person}{Kenneth~O.
  Stanley}.} \bibinfo{year}{2015}\natexlab{}.
\newblock \showarticletitle{Confronting the Challenge of Quality Diversity}. In
  \bibinfo{booktitle}{\emph{Proceedings of the Genetic and Evolutionary
  Computation Conference (GECCO)}}. \bibinfo{publisher}{ACM},
  \bibinfo{address}{New York, NY, USA}, \bibinfo{pages}{967--974}.
\newblock


\bibitem[\protect\citeauthoryear{Runarsson and Yao}{Runarsson and Yao}{2005}]%
        {Runarsson2005}
\bibfield{author}{\bibinfo{person}{T.~P. Runarsson} {and} \bibinfo{person}{Xin
  Yao}.} \bibinfo{year}{2005}\natexlab{}.
\newblock \showarticletitle{Search biases in constrained evolutionary
  optimization}.
\newblock \bibinfo{journal}{\emph{IEEE Transactions on Systems, Man, and
  Cybernetics, Part C (Applications and Reviews)}} \bibinfo{volume}{35},
  \bibinfo{number}{2} (\bibinfo{year}{2005}), \bibinfo{pages}{233--243}.
\newblock


\bibitem[\protect\citeauthoryear{Samuelsen and Glette}{Samuelsen and
  Glette}{2018}]%
        {samuelsen2018multi}
\bibfield{author}{\bibinfo{person}{Eivind Samuelsen} {and}
  \bibinfo{person}{Kyrre Glette}.} \bibinfo{year}{2018}\natexlab{}.
\newblock \showarticletitle{Multi-objective Analysis of MAP-Elites
  Performance}.
\newblock \bibinfo{journal}{\emph{CoRR}}  \bibinfo{volume}{abs/1803.05174}
  (\bibinfo{year}{2018}), \bibinfo{pages}{1--8}.
\newblock
\showeprint[arxiv]{1803.05174}


\bibitem[\protect\citeauthoryear{Shahrzad, Fink, and Miikkulainen}{Shahrzad
  et~al\mbox{.}}{2018}]%
        {shahrzad2018enhanced}
\bibfield{author}{\bibinfo{person}{Hormoz Shahrzad}, \bibinfo{person}{Daniel
  Fink}, {and} \bibinfo{person}{Risto Miikkulainen}.}
  \bibinfo{year}{2018}\natexlab{}.
\newblock \showarticletitle{Enhanced Optimization with Composite Objectives and
  Novelty Selection}.
\newblock \bibinfo{journal}{\emph{CoRR}}  \bibinfo{volume}{abs/1803.03744}
  (\bibinfo{year}{2018}), \bibinfo{pages}{1--7}.
\newblock
\showeprint[arxiv]{1803.03744}


\bibitem[\protect\citeauthoryear{Surry and Radcliffe}{Surry and
  Radcliffe}{1997}]%
        {surry1997comoga}
\bibfield{author}{\bibinfo{person}{Patrick~D Surry} {and}
  \bibinfo{person}{Nicholas~J Radcliffe}.} \bibinfo{year}{1997}\natexlab{}.
\newblock \showarticletitle{The COMOGA method: constrained optimisation by
  multi-objective genetic algorithms}.
\newblock \bibinfo{journal}{\emph{Control and Cybernetics}}
  \bibinfo{volume}{26} (\bibinfo{year}{1997}), \bibinfo{pages}{391--412}.
\newblock


\bibitem[\protect\citeauthoryear{Takahama and Sakai}{Takahama and
  Sakai}{2005}]%
        {takahama2005constrained}
\bibfield{author}{\bibinfo{person}{Tetsuyuki Takahama} {and}
  \bibinfo{person}{Setsuko Sakai}.} \bibinfo{year}{2005}\natexlab{}.
\newblock \showarticletitle{Constrained optimization by $\varepsilon$
  constrained particle swarm optimizer with $\varepsilon$-level control}.
\newblock In \bibinfo{booktitle}{\emph{Soft computing as transdisciplinary
  science and technology}}. \bibinfo{publisher}{Springer},
  \bibinfo{address}{Berlin, Heidelberg}, \bibinfo{pages}{1019--1029}.
\newblock


\bibitem[\protect\citeauthoryear{Takahama and Sakai}{Takahama and
  Sakai}{2010}]%
        {Takahama10eDEag}
\bibfield{author}{\bibinfo{person}{Tetsuyuki Takahama} {and}
  \bibinfo{person}{Setsuko Sakai}.} \bibinfo{year}{2010}\natexlab{}.
\newblock \showarticletitle{Constrained optimization by the $\varepsilon$
  constrained differential evolution with an archive and gradient-based
  mutation}. In \bibinfo{booktitle}{\emph{Congress on Evolutionary Computation
  (CEC)}}. \bibinfo{publisher}{IEEE}, \bibinfo{address}{New York},
  \bibinfo{pages}{1--9}.
\newblock


\bibitem[\protect\citeauthoryear{Urquhart and Hart}{Urquhart and Hart}{2018}]%
        {Urquhart2018WSRP}
\bibfield{author}{\bibinfo{person}{Neil Urquhart} {and} \bibinfo{person}{Emma
  Hart}.} \bibinfo{year}{2018}\natexlab{}.
\newblock \showarticletitle{Optimisation and Illumination of a Real-world
  Workforce Scheduling and Routing Application via Map-Elites}.
\newblock \bibinfo{journal}{\emph{CoRR}}  \bibinfo{volume}{abs/1805.11555}
  (\bibinfo{year}{2018}), \bibinfo{pages}{1--13}.
\newblock
\showeprint[arxiv]{1805.11555}


\bibitem[\protect\citeauthoryear{Vassiliades, Chatzilygeroudis, and
  Mouret}{Vassiliades et~al\mbox{.}}{2017}]%
        {Vassiliades2017}
\bibfield{author}{\bibinfo{person}{Vassilis Vassiliades},
  \bibinfo{person}{Konstantinos Chatzilygeroudis}, {and}
  \bibinfo{person}{Jean-Baptiste Mouret}.} \bibinfo{year}{2017}\natexlab{}.
\newblock \showarticletitle{A Comparison of Illumination Algorithms in
  Unbounded Spaces}. In \bibinfo{booktitle}{\emph{Proceedings of the Genetic
  and Evolutionary Computation Conference (GECCO) Companion}}.
  \bibinfo{publisher}{ACM}, \bibinfo{address}{New York, NY, USA},
  \bibinfo{pages}{1578--1581}.
\newblock


\bibitem[\protect\citeauthoryear{Vassiliades, Chatzilygeroudis, and
  Mouret}{Vassiliades et~al\mbox{.}}{2018}]%
        {Vassiliades2018CVT}
\bibfield{author}{\bibinfo{person}{Vassilis Vassiliades},
  \bibinfo{person}{Konstantinos Chatzilygeroudis}, {and}
  \bibinfo{person}{Jean-Baptiste Mouret}.} \bibinfo{year}{2018}\natexlab{}.
\newblock \showarticletitle{Using Centroidal Voronoi Tessellations to Scale Up
  the Multidimensional Archive of Phenotypic Elites Algorithm}.
\newblock \bibinfo{journal}{\emph{IEEE Transactions on Evolutionary
  Computation}} \bibinfo{volume}{22}, \bibinfo{number}{4}
  (\bibinfo{year}{2018}), \bibinfo{pages}{623--630}.
\newblock
\showISSN{1089-778X}


\bibitem[\protect\citeauthoryear{Vassiliades and Mouret}{Vassiliades and
  Mouret}{2018}]%
        {Vassiliades2018}
\bibfield{author}{\bibinfo{person}{Vassiiis Vassiliades} {and}
  \bibinfo{person}{Jean-Baptiste Mouret}.} \bibinfo{year}{2018}\natexlab{}.
\newblock \showarticletitle{Discovering the Elite Hypervolume by Leveraging
  Interspecies Correlation}. In \bibinfo{booktitle}{\emph{Proceedings of the
  Genetic and Evolutionary Computation Conference (GECCO)}}.
  \bibinfo{publisher}{ACM}, \bibinfo{address}{New York, NY, USA},
  \bibinfo{pages}{149--156}.
\newblock


\bibitem[\protect\citeauthoryear{Venkatraman and Yen}{Venkatraman and
  Yen}{2005}]%
        {Venkatraman2005}
\bibfield{author}{\bibinfo{person}{Sangameswar Venkatraman} {and}
  \bibinfo{person}{Gary~G. Yen}.} \bibinfo{year}{2005}\natexlab{}.
\newblock \showarticletitle{A generic framework for constrained optimization
  using genetic algorithms}.
\newblock \bibinfo{journal}{\emph{IEEE Transactions on Evolutionary
  Computation}} \bibinfo{volume}{9}, \bibinfo{number}{4}
  (\bibinfo{year}{2005}), \bibinfo{pages}{424--435}.
\newblock
\showISSN{1089-778X}


\bibitem[\protect\citeauthoryear{Wang, Yin, Yang, and Sun}{Wang
  et~al\mbox{.}}{2018}]%
        {wangGlobalLocalSurrogate2018}
\bibfield{author}{\bibinfo{person}{Yong Wang}, \bibinfo{person}{Da-Qing Yin},
  \bibinfo{person}{Shengxiang Yang}, {and} \bibinfo{person}{Guangyong Sun}.}
  \bibinfo{year}{2018}\natexlab{}.
\newblock \showarticletitle{Global and Local Surrogate-Assisted Differential
  Evolution for Expensive Constrained Optimization Problems With Inequality
  Constraints}.
\newblock \bibinfo{journal}{\emph{IEEE Transactions on Cybernetics}}
  \bibinfo{volume}{1}, \bibinfo{number}{1} (\bibinfo{year}{2018}),
  \bibinfo{pages}{1--15}.
\newblock
\showISSN{2168-2267}


\bibitem[\protect\citeauthoryear{Zhou, Li, He, and Kang}{Zhou
  et~al\mbox{.}}{2003}]%
        {Zhou2003}
\bibfield{author}{\bibinfo{person}{Yuren Zhou}, \bibinfo{person}{Yuanxing Li},
  \bibinfo{person}{Jun He}, {and} \bibinfo{person}{Lishan Kang}.}
  \bibinfo{year}{2003}\natexlab{}.
\newblock \showarticletitle{Multi-objective and MGG evolutionary algorithm for
  constrained optimization}. In \bibinfo{booktitle}{\emph{Congress on
  Evolutionary Computation (CEC)}}. \bibinfo{publisher}{IEEE},
  \bibinfo{address}{New York}, \bibinfo{pages}{1--5}.
\newblock


\end{thebibliography}

% ------------------------------------------------------------

%\InputIfFileExists{tables/bak/10D_sigma01_no_xover.tex}{}{}
%\InputIfFileExists{tables/bak/10D_sigma05_no_xover.tex}{}{}
%\InputIfFileExists{tables/bak/10D_sigma01_xover.tex}{}{}
%\InputIfFileExists{tables/bak/30D_sigma01_no_xover.tex}{}{}
%\InputIfFileExists{tables/bak/30D_sigma05_no_xover.tex}{}{}
%\InputIfFileExists{tables/bak/30D_sigma01_xover.tex}{}{}

\footnotesize
\begin{table*}[t]
  \caption{Numerical results obtained from 25 independent of MAP-Elites (Configuration 1) on 10D problems}
  \label{tab:10D_standard_cec}
  \begin{tabular}{c|ccccccccc}
\toprule
Function &Best &Worst &Median &$c$ &$\bar{v}$ &Mean &Std & FR\\
\midrule
C01 &-5.900e-01(0) &-3.500e-01(0) &-4.400e-01(0) &(0, 0, 0) &0 &-4.600e-01 &7.000e-02 &1.0\\
C02 &8.900e-01(0) &2.860e+00(2) &1.290e+00(0) &(0, 0, 0) &0 &1.780e+00 &1.870e+00 &0.52\\
C03 &1.254e+14(1) &9.654e+14(1) &5.118e+14(1) &(0, 0, 1) &2.477e+06 &5.208e+14 &2.730e+14 &0.0\\
C04 &-1.512e+01(4) &3.930e+01(4) &1.646e+01(3) &(0, 0, 3) &7.993e+03 &1.487e+01 &1.249e+01 &0.0\\
C05 &-1.795e+02(2) &3.301e+02(2) &8.052e+01(2) &(1, 0, 1) &-8.064e+01 &6.719e+01 &1.716e+02 &0.0\\
C06 &-1.222e+02(2) &5.043e+02(1) &1.705e+02(1) &(0, 0, 1) &-1.300e+02 &1.885e+02 &1.669e+02 &0.0\\
C07 &1.761e+09(0) &3.347e+11(0) &7.117e+09(0) &(0, 0, 0) &0 &6.854e+10 &9.030e+10 &1.0\\
C08 &2.610e+04(0) &1.504e+08(0) &3.490e+06(0) &(0, 0, 0) &0 &1.690e+07 &3.667e+07 &1.0\\
C09 &2.752e+13(0) &2.327e+15(1) &4.526e+13(1) &(0, 1, 0) &-2.950e+01 &3.607e+14 &6.492e+14 &0.12\\
C10 &7.187e+12(1) &8.964e+14(1) &3.631e+13(1) &(0, 0, 1) &-6.560e+02 &1.805e+14 &3.008e+14 &0.0\\
C11 &-5.486e+01(1) &-3.766e+01(1) &-4.441e+01(1) &(0, 0, 1) &2.183e+12 &-4.561e+01 &4.870e+00 &0.0\\
C12 &-5.327e+03(2) &1.704e+03(2) &-3.904e+03(2) &(0, 0, 2) &1.501e+12 &-3.008e+03 &2.173e+03 &0.0\\
C13 &-3.702e+02(2) &-1.721e+02(2) &-2.804e+02(2) &(0, 0, 1) &5.853e+02 &-2.823e+02 &5.116e+01 &0.0\\
C14 &2.141e+14(0) &4.448e+15(0) &1.304e+15(0) &(0, 0, 0) &0 &1.555e+15 &1.018e+15 &1.0\\
C15 &9.018e+12(0) &1.041e+15(0) &4.145e+14(0) &(0, 0, 0) &0 &4.449e+14 &2.936e+14 &1.0\\
C16 &4.900e-01(0) &1.100e+00(2) &6.500e-01(4) &(0, 1, 3) &-4.585e+06 &6.200e-01 &3.000e-01 &0.16\\
C17 &1.310e+00(0) &4.958e+02(1) &1.784e+01(1) &(0, 1, 0) &-8.500e+05 &6.456e+01 &1.025e+02 &0.4\\
C18 &3.365e+01(0) &3.542e+03(1) &5.894e+02(0) &(0, 0, 0) &0 &1.478e+03 &2.594e+03 &0.88\\
\bottomrule
\end{tabular}
\end{table*}
\begin{table*}[t]
  \caption{Numerical results obtained from 25 independent of MAP-Elites (Configuration 2) on 10D problems}
  \label{tab:10D_mutation_cec}
  \begin{tabular}{c|ccccccccc}
\toprule
Function &Best &Worst &Median &$c$ &$\bar{v}$ &Mean &Std & FR\\
\midrule
C01 &-6.700e-01(0) &-4.300e-01(0) &-4.800e-01(0) &(0, 0, 0) &0 &-5.100e-01 &6.000e-02 &1.0\\
C02 &1.700e+00(0) &4.980e+00(1) &2.500e+00(2) &(1, 1, 0) &-1.046e+01 &2.230e+00 &1.930e+00 &0.36\\
C03 &4.516e+13(1) &7.934e+14(1) &4.101e+14(1) &(0, 0, 1) &2.642e+06 &3.827e+14 &1.928e+14 &0.0\\
C04 &-1.654e+01(4) &3.941e+01(4) &1.958e+01(4) &(1, 0, 3) &3.074e+03 &1.224e+01 &1.789e+01 &0.0\\
C05 &-3.516e+02(2) &5.255e+02(2) &-9.369e+01(2) &(1, 1, 0) &-6.720e+01 &-3.351e+01 &2.121e+02 &0.0\\
C06 &-2.519e+02(2) &5.891e+02(2) &-1.068e+02(2) &(0, 1, 1) &1.434e+01 &-2.237e+01 &2.178e+02 &0.0\\
C07 &4.191e+08(0) &2.371e+11(0) &4.222e+09(0) &(0, 0, 0) &0 &2.836e+10 &5.798e+10 &1.0\\
C08 &8.302e+05(0) &2.420e+08(0) &1.664e+07(0) &(0, 0, 0) &0 &4.818e+07 &6.160e+07 &1.0\\
C09 &8.585e+11(1) &7.835e+13(1) &1.933e+13(1) &(1, 0, 0) &-1.388e+02 &2.804e+13 &2.302e+13 &0.0\\
C10 &5.666e+12(0) &3.540e+14(1) &1.854e+13(1) &(1, 0, 0) &-2.489e+02 &5.046e+13 &7.657e+13 &0.16\\
C11 &-5.366e+01(1) &-3.885e+01(1) &-4.573e+01(1) &(0, 0, 1) &1.894e+12 &-4.589e+01 &3.750e+00 &0.0\\
C12 &-6.849e+03(2) &5.433e+02(2) &-4.272e+03(2) &(0, 0, 2) &7.650e+11 &-3.843e+03 &1.779e+03 &0.0\\
C13 &-3.849e+02(2) &-4.087e+01(2) &-3.048e+02(2) &(1, 0, 1) &4.854e+02 &-2.916e+02 &7.552e+01 &0.0\\
C14 &1.004e+07(0) &9.089e+13(0) &8.365e+08(0) &(0, 0, 0) &0 &3.681e+12 &1.780e+13 &1.0\\
C15 &3.640e+06(0) &1.594e+14(0) &4.011e+10(0) &(0, 0, 0) &0 &7.150e+12 &3.112e+13 &1.0\\
C16 &1.060e+00(0) &1.080e+00(4) &9.900e-01(4) &(1, 1, 2) &-5.208e+07 &8.700e-01 &2.300e-01 &0.08\\
C17 &9.251e+01(0) &6.959e+02(3) &3.041e+01(3) &(0, 2, 1) &-3.098e+08 &2.108e+02 &2.661e+02 &0.36\\
C18 &2.253e+02(0) &2.242e+04(1) &8.008e+03(0) &(0, 0, 0) &0 &9.665e+03 &8.152e+03 &0.8\\
\bottomrule
\end{tabular}
\end{table*}
\begin{table*}[t]
  \caption{Numerical results obtained from 25 independent of MAP-Elites (Configuration 3) on 10D problems}
  \label{tab:10D_crossover_cec}
  \begin{tabular}{c|ccccccccc}
\toprule
Function &Best &Worst &Median &$c$ &$\bar{v}$ &Mean &Std & FR\\
\midrule
C01 &-8.300e-01(0) &-4.500e-01(0) &-5.500e-01(0) &(0, 0, 0) &0 &-5.700e-01 &1.000e-01 &1.0\\
C02 &8.000e-01(0) &3.050e+00(2) &2.870e+00(2) &(1, 1, 0) &-8.750e+00 &1.850e+00 &2.230e+00 &0.44\\
C03 &1.154e+14(1) &9.250e+14(1) &4.362e+14(1) &(0, 0, 1) &1.310e+06 &5.005e+14 &2.213e+14 &0.0\\
C04 &-4.441e+01(4) &5.003e+01(4) &3.800e-01(4) &(0, 2, 2) &9.570e+01 &1.270e+00 &2.920e+01 &0.0\\
C05 &-4.801e+02(2) &1.527e+02(1) &-2.316e+02(2) &(0, 1, 1) &1.188e+01 &-2.181e+02 &1.804e+02 &0.0\\
C06 &-4.275e+02(2) &4.210e+02(2) &-1.927e+02(1) &(0, 0, 1) &-1.397e+02 &-1.363e+02 &2.086e+02 &0.0\\
C07 &6.152e+03(0) &5.908e+07(0) &3.171e+05(0) &(0, 0, 0) &0 &4.070e+06 &1.251e+07 &1.0\\
C08 &4.550e+03(0) &1.391e+06(0) &1.518e+05(0) &(0, 0, 0) &0 &4.183e+05 &4.949e+05 &1.0\\
C09 &1.690e+07(0) &1.600e+15(1) &8.227e+07(1) &(0, 1, 0) &-6.704e+01 &1.612e+14 &4.344e+14 &0.2\\
C10 &9.235e+06(0) &7.494e+14(1) &7.169e+07(1) &(1, 0, 0) &-3.527e+01 &3.833e+13 &1.490e+14 &0.44\\
C11 &-5.242e+01(1) &-3.835e+01(1) &-4.532e+01(1) &(0, 0, 1) &3.883e+12 &-4.474e+01 &3.120e+00 &0.0\\
C12 &-7.742e+03(2) &3.236e+02(2) &-5.923e+03(2) &(1, 0, 1) &1.573e+12 &-5.795e+03 &1.603e+03 &0.0\\
C13 &-4.474e+02(2) &-2.925e+02(2) &-3.952e+02(2) &(0, 1, 1) &8.816e+02 &-3.900e+02 &3.742e+01 &0.0\\
C14 &1.033e+05(0) &3.280e+14(0) &1.724e+09(0) &(0, 0, 0) &0 &3.958e+13 &8.132e+13 &1.0\\
C15 &4.980e+03(0) &4.089e+11(0) &9.060e+07(0) &(0, 0, 0) &0 &1.863e+10 &8.010e+10 &1.0\\
C16 &1.020e+00(0) &1.100e+00(4) &2.900e-01(3) &(2, 1, 0) &-3.723e+05 &5.100e-01 &3.800e-01 &0.12\\
C17 &3.610e+00(0) &1.560e+01(3) &1.123e+01(3) &(2, 1, 0) &-8.454e+05 &6.054e+01 &1.196e+02 &0.52\\
C18 &7.190e+00(0) &1.480e+01(2) &5.091e+01(0) &(0, 0, 0) &0 &1.078e+02 &1.165e+02 &0.92\\
\bottomrule
\end{tabular}
\end{table*}
\begin{table*}[t]
  \caption{Numerical results obtained from 25 independent of MAP-Elites (Configuration 1) on 30D problems}
  \label{tab:30D_standard_cec}
  \begin{tabular}{c|ccccccccc}
\toprule
Function &Best &Worst &Median &$c$ &$\bar{v}$ &Mean &Std & FR\\
\midrule
C01 &-2.800e-01(0) &-2.000e-01(0) &-2.200e-01(0) &(0, 0, 0) &0 &-2.300e-01 &2.000e-02 &1.0\\
C02 &3.880e+00(0) &4.320e+00(2) &3.760e+00(1) &(1, 0, 0) &-2.570e+00 &3.210e+00 &1.480e+00 &0.36\\
C03 &6.664e+15(1) &1.887e+16(1) &1.472e+16(1) &(0, 0, 1) &1.798e+07 &1.463e+16 &3.100e+15 &0.0\\
C04 &1.869e+01(4) &4.507e+01(4) &3.239e+01(4) &(1, 1, 2) &1.527e+06 &3.235e+01 &5.850e+00 &0.0\\
C05 &1.739e+02(2) &4.920e+02(1) &2.784e+02(1) &(0, 1, 0) &-1.224e+01 &3.036e+02 &8.686e+01 &0.0\\
C06 &2.237e+02(2) &5.782e+02(1) &3.076e+02(2) &(0, 1, 1) &5.920e+00 &3.545e+02 &1.043e+02 &0.0\\
C07 &3.226e+10(0) &6.284e+12(0) &9.281e+11(0) &(0, 0, 0) &0 &1.328e+12 &1.386e+12 &1.0\\
C08 &6.346e+07(0) &6.746e+10(0) &1.811e+09(0) &(0, 0, 0) &0 &7.092e+09 &1.399e+10 &1.0\\
C09 &7.159e+14(0) &3.818e+15(1) &8.511e+14(1) &(0, 0, 1) &-1.028e+03 &1.323e+15 &9.556e+14 &0.2\\
C10 &6.189e+14(0) &3.359e+15(1) &7.702e+14(1) &(1, 0, 0) &-1.984e+02 &9.456e+14 &7.034e+14 &0.16\\
C11 &-3.289e+01(1) &-2.454e+01(1) &-2.802e+01(1) &(0, 0, 1) &5.839e+12 &-2.833e+01 &2.180e+00 &0.0\\
C12 &-1.363e+04(2) &2.778e+03(2) &-8.056e+03(2) &(0, 1, 1) &3.562e+12 &-7.862e+03 &3.439e+03 &0.0\\
C13 &-3.129e+02(2) &-1.248e+01(2) &-2.373e+02(2) &(0, 1, 1) &5.074e+02 &-2.253e+02 &5.712e+01 &0.0\\
C14 &6.971e+15(0) &2.152e+16(0) &1.160e+16(0) &(0, 0, 0) &0 &1.203e+16 &3.312e+15 &1.0\\
C15 &1.754e+15(0) &2.898e+16(0) &6.434e+15(0) &(0, 0, 0) &0 &7.380e+15 &5.067e+15 &1.0\\
C16 &9.100e-01(0) &1.050e+00(3) &8.500e-01(4) &(1, 2, 1) &-1.920e+19 &8.200e-01 &1.700e-01 &0.08\\
C17 &6.822e+01(0) &1.316e+03(2) &8.772e+02(0) &(0, 0, 0) &0 &9.761e+02 &8.256e+02 &0.64\\
C18 &1.228e+03(0) &1.276e+04(0) &3.849e+03(0) &(0, 0, 0) &0 &4.380e+03 &2.752e+03 &1.0\\
\bottomrule
\end{tabular}
\end{table*}
\begin{table*}[t]
  \caption{Numerical results obtained from 25 independent of MAP-Elites (Configuration 2) on 30D problems}
  \label{tab:30D_mutation_cec}
  \begin{tabular}{c|ccccccccc}
\toprule
Function &Best &Worst &Median &$c$ &$\bar{v}$ &Mean &Std & FR\\
\midrule
C01 &-2.600e-01(0) &-2.100e-01(0) &-2.300e-01(0) &(0, 0, 0) &0 &-2.400e-01 &1.000e-02 &1.0\\
C02 &3.510e+00(0) &4.940e+00(2) &3.760e+00(2) &(0, 2, 0) &-5.310e+00 &3.440e+00 &1.030e+00 &0.24\\
C03 &6.155e+15(1) &1.970e+16(1) &1.396e+16(1) &(0, 0, 1) &1.316e+07 &1.378e+16 &3.582e+15 &0.0\\
C04 &1.892e+01(4) &4.564e+01(4) &3.525e+01(4) &(0, 1, 3) &1.317e+06 &3.429e+01 &6.280e+00 &0.0\\
C05 &-2.140e+00(2) &5.772e+02(1) &2.484e+02(2) &(0, 1, 1) &-2.131e+01 &2.570e+02 &1.174e+02 &0.0\\
C06 &1.318e+02(2) &5.961e+02(1) &2.776e+02(1) &(0, 0, 1) &-1.296e+02 &3.380e+02 &1.436e+02 &0.0\\
C07 &6.860e+10(0) &6.957e+12(0) &2.374e+12(0) &(0, 0, 0) &0 &2.560e+12 &1.740e+12 &1.0\\
C08 &4.992e+08(0) &1.523e+12(0) &6.357e+10(0) &(0, 0, 0) &0 &1.370e+11 &2.923e+11 &1.0\\
C09 &1.715e+15(0) &1.520e+15(1) &7.103e+14(1) &(1, 0, 0) &4.026e+02 &7.634e+14 &4.300e+14 &0.08\\
C10 &1.375e+14(1) &1.511e+15(1) &7.229e+14(1) &(0, 0, 1) &-5.762e+02 &7.349e+14 &3.095e+14 &0.0\\
C11 &-3.339e+01(1) &-2.650e+01(1) &-2.925e+01(1) &(0, 0, 1) &5.776e+12 &-2.991e+01 &2.060e+00 &0.0\\
C12 &-1.146e+04(2) &-4.734e+01(2) &-7.854e+03(2) &(0, 1, 1) &3.687e+12 &-6.797e+03 &3.248e+03 &0.0\\
C13 &-3.302e+02(2) &2.444e+01(2) &-2.143e+02(2) &(1, 0, 1) &5.036e+02 &-1.960e+02 &8.269e+01 &0.0\\
C14 &2.752e+10(0) &4.214e+13(0) &3.568e+12(0) &(0, 0, 0) &0 &7.098e+12 &1.001e+13 &1.0\\
C15 &1.921e+09(0) &2.006e+15(0) &1.110e+13(0) &(0, 0, 0) &0 &1.083e+14 &3.901e+14 &1.0\\
C16 &1.030e+00(4) &1.300e+00(2) &1.110e+00(4) &(0, 2, 2) &-6.197e+17 &1.130e+00 &8.000e-02 &0.0\\
C17 &1.110e+03(0) &2.864e+03(3) &1.417e+03(0) &(0, 0, 0) &0 &1.539e+03 &7.542e+02 &0.44\\
C18 &5.037e+03(0) &4.064e+04(0) &3.043e+04(0) &(0, 0, 0) &0 &2.847e+04 &9.539e+03 &1.0\\
\bottomrule
\end{tabular}
\end{table*}
\begin{table*}[t]
  \caption{Numerical results obtained from 25 independent of MAP-Elites (Configuration 3) on 30D problems}
  \label{tab:30D_crossover_cec}
  \begin{tabular}{c|ccccccccc}
\toprule
Function &Best &Worst &Median &$c$ &$\bar{v}$ &Mean &Std & FR\\
\midrule
C01 &-3.100e-01(0) &-2.300e-01(0) &-2.600e-01(0) &(0, 0, 0) &0 &-2.600e-01 &2.000e-02 &1.0\\
C02 &4.590e+00(0) &5.160e+00(2) &3.150e+00(1) &(1, 0, 0) &-1.940e+00 &2.610e+00 &2.260e+00 &0.32\\
C03 &1.033e+16(1) &2.039e+16(1) &1.521e+16(1) &(0, 0, 1) &9.516e+06 &1.534e+16 &2.809e+15 &0.0\\
C04 &-1.859e+01(4) &4.163e+01(4) &1.518e+01(4) &(0, 2, 2) &3.534e+05 &1.215e+01 &1.744e+01 &0.0\\
C05 &-2.304e+02(1) &5.599e+02(2) &-7.310e+01(1) &(0, 0, 1) &-6.470e+00 &9.960e+00 &1.769e+02 &0.0\\
C06 &-1.678e+02(1) &5.736e+02(1) &-1.477e+01(1) &(0, 0, 1) &-8.221e+01 &7.093e+01 &2.096e+02 &0.0\\
C07 &7.226e+05(0) &2.288e+09(0) &2.058e+07(0) &(0, 0, 0) &0 &1.503e+08 &4.498e+08 &1.0\\
C08 &6.173e+05(0) &3.216e+07(0) &2.674e+06(0) &(0, 0, 0) &0 &6.223e+06 &7.284e+06 &1.0\\
C09 &5.770e+09(0) &6.548e+10(1) &1.463e+10(0) &(0, 0, 0) &0 &2.184e+13 &1.050e+14 &0.48\\
C10 &1.301e+08(0) &1.153e+11(1) &4.820e+09(1) &(1, 0, 0) &-2.706e+01 &4.189e+13 &1.970e+14 &0.48\\
C11 &-3.475e+01(1) &-2.439e+01(1) &-2.780e+01(1) &(0, 0, 1) &5.744e+12 &-2.825e+01 &2.340e+00 &0.0\\
C12 &-1.818e+04(2) &2.716e+03(2) &-1.583e+04(2) &(0, 0, 2) &3.737e+12 &-1.369e+04 &5.767e+03 &0.0\\
C13 &-3.659e+02(2) &-2.891e+02(2) &-3.336e+02(2) &(1, 0, 1) &6.148e+02 &-3.291e+02 &2.011e+01 &0.0\\
C14 &2.949e+06(0) &9.116e+15(0) &3.783e+08(0) &(0, 0, 0) &0 &6.363e+14 &2.134e+15 &1.0\\
C15 &9.592e+05(0) &4.604e+09(0) &1.375e+07(0) &(0, 0, 0) &0 &2.319e+08 &8.981e+08 &1.0\\
C16 &9.600e-01(0) &1.040e+00(4) &8.100e-01(4) &(1, 0, 3) &-1.020e+20 &8.300e-01 &1.400e-01 &0.04\\
C17 &6.458e+01(0) &1.234e+02(3) &8.320e+01(3) &(1, 1, 1) &-1.291e+20 &5.433e+02 &7.307e+02 &0.48\\
C18 &1.570e+02(0) &2.793e+04(0) &4.208e+02(0) &(0, 0, 0) &0 &1.896e+03 &5.424e+03 &1.0\\
\bottomrule
\end{tabular}
\end{table*}

%\InputIfFileExists{tables/10D_eDAag.tex}{}{}
%\InputIfFileExists{tables/30D_eDAag.tex}{}{}

\end{document}